# Exploratory Methods for Relation Discovery in Archival Data

Giagnolini Lucia[a*], M. Daquino[b], F. Mambelli[c], and F. Tomasi[d]

[a]Historical Archive of the University of Bologna, Bologna, Italy, ORCID: 0000-0002-4876-2691;
[b]Digital Humanities Advanced Research Centre (DH.arc), University of Bologna, Bologna, Italy, ORCID: 0000-0002-1113-7550;
[c]Fondazione Federico Zeri, University of Bologna, Bologna, Italy, ORCID: 0000-0002-2341-4375;
[d]Digital Humanities Advanced Research Centre (DH.arc), University of Bologna, Bologna, Italy, ORCID: 0000-0002-6631-8607.

[*]Lucia Giagnolini, lucia.giagnolini@studio.unibo.it

**Abstract**

In this article we propose a holistic approach to discover relations in art historical communities and enrich historians' biographies and archival descriptions with graph patterns relevant to art historiographic enquiry. We use exploratory data analysis to detect patterns, we select features, and we use them to evaluate classification models to predict new relations, to be recommended to archivists during the cataloguing phase. Results show that relations based on biographical information can be addressed with higher precision than relations based on research topics or institutional relations. Deterministic and *a priori* rules present better results than probabilistic methods.

## 1. Introduction[1]

Historiography of art encompasses the legacy of scholarship in art history, focusing on the relations between scholars, museum professionals, art market representatives, and connoisseurs. Methods for performing quantitative analysis relevant to historiography of art are still in their infancy. Among the reasons, the lack of authoritative, complete, structured data sources ready to be used for data analysis is a major obstacle (Fiorucci et al., 2020; Jaskot, 2020). Moreover, when available, such sources often require data integration and cross-linking to discover latent relations and have a complete overview for historiographic purposes (Vander Sande et al., 2018).

ARTchives[2] is a newborn project for crowdsourcing data on art historians' archives. The catalogue collects information on art historical periods, people (including artists and historians), archival collections, and institutions (linked to artists, art historians, and collections). Currently, the pilot project counts around 27 collections produced by 26 historians and held by 6 institutions. Although narrow, the scope covers studies representative of Italian Modern art history.

Data is provided by archivists and historians belonging to holding institutions, and an editorial board supervises the data collection process to ensure data quality. Nonetheless, data is recorded with different levels of insight, and data completeness cannot be ensured for all records. Moreover, only associative relations between art historians and research topics are available in ARTchives, and it is not always clear whether these are meaningful for historiographic research, that is, whether relations refer to the history of scholarship and provide insights on competing approaches to artistic movement or artists. As a result, most of the actual network relations between historians, their working materials, and their subjects of study, remain unveiled or too vague, hampering the objectives of the project in the long term.

The objective of this work is to explore how deterministic and probabilistic methods combined with Linked Open Data can be used to identify additional relations. This can then lead to developing a recommender system to support cataloguers in relation discovery. We propose a holistic approach to characterise and predict relations between historians and between archival collections. We first characterise an art history community according to significant dimensions via Exploratory Data Analysis, and we evaluate classification models to make predictions to be recommended to cataloguers when recording or updating information of an archival collection, therefore speeding up their task and preventing information loss.

In detail, we (1) extract information from ARTchives Linked Open Data and Wikidata, (2) use graph patterns, Natural Language Processing methods, and exploratory data analysis, to detect recurring patterns in art historical communities, (3) ask experts to manually annotate our dataset to create a gold standard, and (4) use several classification methods to extract and predict new relations between historians and between their collections.

In previous work, (Jo and Gebru, 2020) have highlighted the differences between archival practices and Machine Learning community approaches to select, annotate, and supervise data collection. While the former privilege highly supervised data collection practises, the latter may tend to emphasize size and efficiency. We believe that the validity of recommendations in the cultural heritage domain should rely on precision rather than recall, so as to serve high-quality, trustworthy information and replicate archivists' methodology. Therefore, precision is our main reference in the evaluation of classification tasks.

Results show that historians' relations extracted from biographies can be identified with higher precision than topic-based and institution-based relations. Relations between collections are difficult to predict, and again biographical relations provide more insights. Deterministic rules are generally valid, while probabilistic methods present challenges. Our methodology can be applied to any similar archival data source and can effectively support cataloguers in the definition of methods for recommending precise information.

The remainder of the article is the following. In section 2 we provide an overview of data sources for art historiography, we motivate the choice of our gold standard, and we address issues related to relations extraction and classification. In section 3 we provide information on ARTchives dataset, our research objectives, and our approach to detect graph patterns. In section 4 we show results of the manual annotation of ARTchives, the exploratory data analysis, and the evaluation of three classifiers to predict relations. In section 5 we discuss our results, in the light of benefits and limitations. Lastly, we conclude with future perspectives.

## 2. Related Work
*2.1 Data Sources for Historiography of Art*
Nowadays, several digital resources support scholars in art historiography enquiries, such as the Dictionary of Art Historians (Knight, 2011), a catalogue of biographies of scholars and museum professionals, or Kubikat (Ebert-Schifferer, 2005), the bibliographic database of the major German art history research libraries in Florence, Munich and Rome. In recent years, Linked Open Data (LOD) technologies have become a priority for cultural institutions that want to serve high-quality data according to shared vocabularies. International consortia of archives and museums (Caraffa et al.,

2020; Delmas-Glass and Sanderson, 2020) provide a variety of cataloguing data on cultural objects, including bibliographic and historical information. However, different granularities of information are in place. Moreover, relations between art historians are often disregarded in museum data.

Art historical photo archives are privileged sources for documenting the work of notable scholars. For instance, the Federico Zeri's photo archive[3] includes detailed information on his research interests and methods, and his relations with other historians and art market representatives. A significant subset of data is available as Linked Open Data (Daquino et al., 2017). Unfortunately, other archives do not provide open data for reuse, therefore only partial information is available. General-purpose sources, like Wikidata, offer curated data about historians, institutions, and bibliographic sources, which can complement existing catalogues. However, Wikidata rarely describes in detail the contents of historians' collections.

ARTchives is an ongoing project that aims at becoming a specialistic source for historiographic research. Like other crowdsourced datasets, data completeness is not ensured. Whenever applicable, alignment to Wikidata information is provided, thanks to the native catalogue recommending system that suggests candidate matches to cataloguers when entering a new record. Moreover, it allows to record a variety of relationships significant for historiographic research, namely: historians-historians, historians-topics, historians-institutions, historians-collections, collections-topics, collections-institutions. Collections described in ARTchives are representative of the scholarship on Italian Modern art, therefore providing a homogeneous corpus for framing the problem of relation discoverability and recommendation.

*2.2 Relation Discovery*

Relation extraction is a well-known computer-aided task investigated in many application domains (Pawar et al., 2017). Named Entity Recognition (NER) techniques are used to extract real-world entities (e.g. people, institutions, places) from unstructured texts and have become crucial in enriching cultural heritage data (Van Hooland et al., 2015). Supervised methods can leverage labelled data with individual relations between such entities and classifiers can be constructed on the basis of linguistic or semantic features to predict new relations. Among the most used supervised learning methods in the cultural heritage domain there are Logistic regression and Decision Tree (Fiorucci et al., 2020). Also, Naive Bayes is used for text classification purposes in archives (Liu et al., 2017).

However, on the one hand, detecting relations based on semantic features is challenging, since features may not be appropriately represented in the dataset (Fiorucci et al., 2020), and background knowledge provided by humans may be required. On the other hand, probabilistic models are not always necessary to detect relations and deterministic rules (e.g. decision-trees or rules proposed by domain experts) may be sufficient.

In this work we explore the feasibility of deterministic rules and classification methods to identify relation patterns that are not explicitly mentioned in the text but that are likely to happen according to background knowledge provided by domain experts. Methods are tested over the ARTchives dataset in order to define the best recommendation algorithm.

### 3. Materials and Methods

*3.1 The Dataset*

The ARTchives pilot project currently includes around 27 archival collections, produced by 26 historians, held by 6 institutions, representative of the scholarship on Italian Modern art. Historians' floruits span from the end of 19th century to the beginning of 21th century. Detailed descriptions of collections and historians are provided as both free-text descriptions and structured data. Relations involve four main subjects, namely: research topics (such as artists and periods), art historians, archival collections, and institutions.

The ARTchives dataset is available as Linked Open Data (Bizer Heath and Berners-Lee, T., 2011) via a dedicated SPARQL endpoint.[4] Relations are represented as triple patterns <subject> <predicate> <object> in the RDF dataset. Such relations mainly address aspects relevant to historians' biographies and descriptions of collections, which are recorded by cataloguers via a web form. Relations are stored in dedicated Named Graphs (Carroll et al., 2005), so that competing information on same entities (e.g. different biographic information for the same historians) are appropriately addressed with their provenance information (e.g. people responsible for data entry, publication dates).

People and organisations mentioned in the biography and the description of collections are automatically extracted via DBpedia spotlight[5] and with SpaCy[6] to extract relations (historian-historian and historian-institution). Extracted entities are reconciled to Wikidata so as to disambiguate and uniquely identify them. It's worth noting that in ARTchives relations are often vague and associative (e.g. "is subject of") and no additional information is given on the scope of the association. For instance, when describing Aby Warburg's archival collection,[7] cataloguers added Federico Zeri as a relevant subject of the collection. However, it is unclear whether the relation refers to materials of the collection (e.g. documents produced by Zeri), or to other forms of relations (e.g. collaborations, institutional relations, similar studies, influence). Such a piece of information was manually added by cataloguers in a form field dedicated to "subjects" (which integrates NER results), but there is no explanation neither in the biography nor in the description of the collection.

*3.2 Approach and Methodology*

Our approach is graphically represented in Fig. 1, and consists of the following main steps:
- we query and analyse ARTchives dataset to highlight networks peculiar to the art historical community (Section 3.3);
- we expand ARTchives to create a labelled dataset including relations annotated by domain experts (Section 3.4);
- we analyse the expanded dataset to support features selection (Section 4.1);
- we evaluate classification models to predict relations between historians and between collections (4.2);

In detail, we query ARTchives and Wikidata and we produce an initial dataset including relations between historians that were identified as interesting patterns by experts. Experts are art historians and cataloguers of the Federico Zeri Foundation and the Historical Archive of the University of Bologna (BUB-Sezione Archivio Storico). According to experts, analysing shared research topics and/or institutional networks (i.e. education places and workplaces) may lead to the discovery of clusters and relation patterns. We perform a preliminary exploratory analysis to investigate historians' network relations and to validate their assumption. A copy of the ARTchives dataset used in this work is available online (Daquino, 2021), and the preliminary analysis is reproducible via an online Jupyter notebook[8].

Secondly, we expand the dataset of existing relations with new assumed relations. *Existing* relations are those recorded in ARTchives, which are incomplete due to aforementioned data quality issues. *Assumed* relations are those we include on the basis of aforementioned assumptions, that is, all pairs of historians that have (1) at least one research topic in common, (2) one institution in common, and (3) at least one topic and one institution in common. We validate assumed relations via manual annotation, performed by the same experts. Two annotators confirm whether an actual relation occurred or not and whether the relation could be inferred from data already in ARTchives (e.g. a historian is mentioned in the biography of the other historian). The annotated dataset is described in section 3.4 and it is available online (Daquino and Giagnolini, 2021).

We perform Exploratory Data Analysis (EDA) on the expanded dataset in order to identify recurring patterns in true positive and false positive relations between historians, research topics, collections, and institutions. The EDA is performed using standard languages (Python) and well-known libraries for Linked Open Data manipulation (e.g. RDFlib, SPARQLWrapper), NLP (spacy, nltk), data analysis (pandas, numpy) and data visualization (e.g. Pyvis). The analysis is described in section 4.1 and can be reproduced via an online interactive Jupyter notebook.[9]

Results of the EDA over our gold standard dataset are used for feature selection. Features are used to construct and evaluate a classifier that estimates the probability of new relations. We adopt a holistic approach to test several well-known classification methods, namely Logistic regression, Naive Bayes, and Decision Tree. We identified these models based on the classificatory problem at hand, and the possibility to interpret results. We use cross-validation to select the model with the highest precision. The decision to privilege precision over recall and accuracy is pragmatic. Since the creation of new cataloguing records is an expensive and time-consuming task, providing correct - despite fewer - recommendations is fundamental to speed up the review process and to boost cataloguers' confidence in leveraging automatic methods to perform their daily tasks. Providing recommendations with higher recall that are not precise might lead to a costly manual inspection of those links, slowing down the annotation process and wasting annotators' time. The evaluation of models is described in section 4.2 and is available in a dedicated Jupyter notebook.[10]

*3.3 Preliminary Analysis*
In the preliminary exploratory analysis, we investigated relations between art historians' that share (1) research topics (including artists and artistic periods) and (2) the institutions in which they worked or studied. We selected these relations on the basis of suggestions from domain experts.

**Research topics patterns**. In Fig. 2, we illustrate art historians' network based on their research subjects. Scholars are marked as light green dots, and their connections are weighted on the basis of the number of shared research topics (dark green dots). Three main clusters (CL) emerge:
- CL1. Ernst Steinmann, Konrnèl Fabriczy, John Pope-Hennessy, and Everett Fahy. While the scholars share general subjects (e.g. Renaissance), specific subjects include artists like Michelangelo and Botticelli.
- CL2. Richard Krautheimer, Federico Zeri, Stefano Tumidei, Wolfgang Lotz, and Roberto Longhi. All scholars studied Gian Lorenzo Bernini and Caravaggio.

- CL3. Fritz Heinemann, Gustav Ludwig, Leo Steinberg, Luisa Vertova, and Roberto Longhi. Scholars' research focused on Vittore Carpaccio, Titian, and Giovanni Bellini.

According to domain experts, clusters are representative of actual collaborations between the scholars in many cases. Therefore, we assume research topics may influence the likelihood of collaborations.

**Institutions patterns.** Similarly, the analysis of workplaces and education places showed us interesting patterns. In Fig. 3 we show the distribution of institutions related to historians recorded in ARTchives. Well-known clusters emerge around Italy and Germany, as confirmed by scholars (Caraffa, 2011), and in the U.S. east coast.

Historians' networks based on shared institutions (Fig. 4) are moderately dense (0.55). Less known relations emerge, which may integrate information missing in the network based on research topics only. Only in a few cases historians that seem strongly related, e.g. Federico Zeri and Leo Steinberg, have never collaborated on any specific topic. Therefore, we assume that having several institutions in common may not be a necessary condition to determine an actual relation, although it increases the likelihood.

*3.4 Dataset Expansion and Annotation*

Based on domain experts' suggestions, we assume that relations based on shared topics and shared institutions may lead us to discover recurring patterns. To validate our assumption, we expand the original dataset with pairs of historians when the following conditions are respected: historians share one topic (i.e. the research topic is linked to both historians in the ARTchives graph); historians share one institution (i.e. the institution is linked to both historians in the ARTchives graph). Moreover, we assume that evidence of actual relations between historians are recorded in their archival collections. Therefore, we extend the dataset with historian-collection relations when the aforementioned conditions are respected.

In summary, 23 out of 26 historians were taken into account, since (1) some descriptions or biographies do not mention topics or institutions, and (2) some topics or institutions are not mentioned in more than one historian's record, and no pairs can be created for that topic or institution.

The expanded dataset is created automatically. For convenience, we produced two tables, respectively including all pairs of historians sharing research topics (Table *artists_periods*), and all pairs of historians having shared institutions (Table *institutions*). In detail:

- *artists_periods*: each row records a pair of historians and their shared topic. If the two historians share more than one topic, the pair appears in as many rows as the number of shared topics.
- *institutions*: each row records a pair of historians and a shared institution. Again, pairs may appear in several rows if these share multiple institutions.

In Table 1 we provide an overview of columns and methods we used to produce value results. Methods include direct SPARQL queries (graph extraction), experts' annotation (manual annotation), and Named Entity Recognition (NER). When the method includes NER, we also mention the source field (biography or archival description of the collection) wherefrom entities were extracted.

Relations between historians are validated with domain experts of the Federico Zeri Foundation and the Historical Archive of the University of Bologna (BUB-Sezione Archivio Storico), who annotated the columns A4, A7, A8, A10, I4, and I5. Annotators confirmed/rejected whether individual associations were true (i.e. there was an actual collaboration between two historians) or false, filling cells with respectively 1 and 0.

When relations could not be validated due to the lack of documentation, the relation was deemed false (and the cell was automatically filled with 0). In a few cases, uncertain relations have been recorded as 0.5. Due to the limited number of such cases, we replaced 0.5 with 0 in our final analysis.

Lastly, in Table 2 we summarise our desiderata in terms of new relations, and the validation method to create our gold standard dataset. We list relations ("Predicates") that are already available in ARTchives (Status "available") for the sake of completeness, relations that we aim to add (Status "new"), and the validation method we used to verify them. The validation method consists in either automatic extraction from the graph patterns (identified by column #, referencing predicates in the same Table 2) or by manual annotation (referencing the column ID # of Table 1).

For instance, we assume two historians are likely to have interacted with each other (Table 2, row #5) if they have at least one research topic in common. Research topics are extracted from historians' biographies and are (partially) available as predicates in the dataset (#2). In the event an explicit relation between historians (#4) is missing in the dataset, cataloguers annotate the relation (#A4) to confirm or reject the relation (#5), and the scope of the interaction (#6). Interactions include any form of direct contact between two people, such as co-authored publications and correspondence.

As an example, consider the relation between Aby Warburg and Ernst Steinmann. From ARTchives records we know that both historians have studied Renaissance (#2). However, there is no evidence in their biographies that the two persons are relevant to each other (#4). Cataloguers confirmed an interaction between the two of them existed (#5) despite it was not recorded, and it was indeed on the topic at hand (#6). Moreover, cataloguers confirmed that Warburg's archive includes documents relevant to Steinmann (#9).

Notice that, when a new relation is proposed between entities for whom a vague relation already exists, the new relation is meant to be a specialization of the previous one (e.g. *subject* and *interacted with*). Inverse relations are not addressed for the sake of brevity. We consider only relations between historians that are both recorded in ARTchives.

In summary, experts identified research topics and institutions as potential factors to detect relation patterns in art historical communities. We validated experts' assumptions via preliminary exploratory analysis, we expanded our dataset to include all pairs of historian-historian relations based on aforementioned patterns, and we asked expert annotators to label expanded relations. Results of the annotation are shown in the Section 4.1, wherein we describe the support of valid and invalid relations.

## 4. Results
### 4.1 Exploratory Data Analysis
The objective of the exploratory data analysis on the expanded dataset is to detect recurring patterns in art historical communities with respect to valid relations between historians and between collections. Valid relations (i.e. the two historians actually collaborated) and invalid relations (i.e. the relation between historians did not exist, despite they shared topics or institutions) are evaluated by annotators, who have the background knowledge necessary to confirm/reject a relation. We characterise valid (true positive) and invalid (false positive) relations with respect to shared topics, shared institutions, and both shared institutions and topics. In particular, we are interested in two patterns, namely:

- The one-to-one relation between two historians (*<historian1> <historian2>*), regardless of the number of topics or institutions in common. In the following paragraphs, we refer to these elements as "unique pairs", i.e. bi-directional direct links between historians.
- The relation between two historians and the specific topic or institution in common (*<historian1> <historian2> <subject>)*. When historians share more than one subject, the pair is considered as many times as the number of common subjects that are relevant. We refer to these patterns as "total relations", i.e. non unique pairs of historians appearing in unique triplets.

Results are summarised in Table 3 and Table 4.

**Relations between historians**. Table 3 summarises results of the exploratory analysis on the relations between historians. The expanded dataset (see Table 1 *artists_periods*) includes 332 relations (results shown in Table 3, row 1). The complete set of relations addresses 23 unique historians, 24 unique topics, and 173 unique pairs of historians (Table 3, row 1, col. *2*). 48% (162) of the total number of relations are valid (Table 3, row *1*, col. *3*). Valid relations relate 23 unique historians in 71 unique pairs of art historians (Table 3, row *1*, col. 4).

It's worth noting that when a historian is mentioned in another historian's biography (*artists_periods*, col. A5), the relation is always valid 100%. Likewise, relations between historians that are (1) mutually mentioned in their respective biographies (*artists_periods*, A6), or (2) mentioned in at least one archival description (*artists_periods*, A9, A11), are 100% valid. Therefore, we are interested in those relations that are not recorded in any way in ARTchives (i.e. column *5. Valid relations not recorded in ARTchives*). Currently, 23% of valid relations (38 out of 162) are recorded in ARTchives in at least one biography, highlighting 67% undisclosed valid relations (124 out of 162) that could be integrated in the catalogue (Table 3, row 1, col. 5).

Among all valid relations (52 out of 162), 32% are based on the research topic at hand (Table 3, col. 6). Such relations address 18 unique art historians in 28 unique pairs, and 12 unique topics.

60 relations between art historians have at least one institution in common (see Table 1. *institutions,* results shown in Table 3, row *2*). Relations address 23 unique art historians in 49 unique pairs, and 19 unique institutions (Table 3, row 2, col. 2). Among the total relations (60), 65% (39) relations are valid (Table 3, row 2, col. 3), which address 21 unique art historians combined in 33 unique pairs of historians (Table 3, row 2, col. 4). 26% relations (10) are already recorded in at least one biography, leaving us with 74% (29) unknown relations that could be integrated (Table 3, row 2, col. 5).

Lastly, we merged the two tables - a*rtists_periods* and *institutions* - into a new table that includes only relations between historians that have at least one institution and one research topic in common. We obtain 173 relations (Table 3, row 3, col. 1) among which 71 relations are valid. Relations address 32 unique pairs of historians, among which 68.5% (22) unique pairs are valid. Out of valid relations, only 20% (14) have been already recorded in ARTchives biographies. Therefore 80% (57) of relations are missing (Table 3, row 3, col. 5).

**Relations between collections**. Table 4 summarises results of the exploratory analysis on relations between historians and collections.

Out of 162 valid relations (see Table 3, row 1, col. 3), only 19% (31) relations between collections and historians are also valid (Table 4, row 1, col. 1). Among these, 61% (19) relations between collections and between historians are missing (Table 4, row 1, col. 2). When historians share at least one topic and one institution, a collection is relevant to the other historian 13 times (Table 4, row 2, col. 1). Among these, 4 relations are not recorded in biographies, (Table 4, row 2, col. 2).

In summary, results confirm that people who collaborated may be recorded in biographies and archival descriptions, and in this case the relation is always valid (i.e. no other personal, unrelated relations are usually recorded in biographies). In these cases, we expect a (deterministic) rule can be defined *a priori* for recommendation purposes. However, 68% of relations that are based on research topics, 74% of institutional relations (and 80% of relations based on both) are still not recorded, which becomes our target for classification tasks. The low support of relations based on a specific topic at hand (32%) does not allow us to address this information in classification tasks. Therefore, we limit our research to bidirectional links between historians, using the pattern "interacted with".

When a collection includes materials relevant to a historian, the relation between historians is always valid. This leads us to define another *a priori* rule for recommendation purposes. On the contrary, when relations between historians are valid, materials relevant to each other are not always preserved in their collections (i.e. a historian is not relevant to another historian's collection). Unfortunately, when analysing relations between collections and historians, we are not able to identify the direction of the link, i.e. which collection includes material of the other historian, since the premises (sharing topics or institutions) are symmetric. Therefore, we limit our research to bidirectional links between historians' collections using the pattern "is related to", and again between historians using the pattern "interacted with".

*4.2 Evaluation of classification models for historiographic relations*
Results of the EDA allow us to identify relations between historians and relations between collections as potential classification tasks. Due to limitations in the support of features, we are able to evaluate the following triple patterns: *<Historian> <interacted with> <Historian>* and *<Collection> <is related to> <Collection>*.

In Table 5 and Table 6 we summarize the results of the evaluation of classification models with respect to the two aforementioned tasks. For each model (lr=logistic regression; nb=naive bayes; dt=decision tree) we include features (column names), mean precision (p), precision on true positive results (p(1)), recall of true positive values (r(1)), and accuracy (a). We judge results based on p(1) and r(1) mainly.
Results of the model that performed better are highlighted in bold.
Results can be summarised as follows:
- When a historian is recorded in another historian's biography (bio) or as a related person of her/his archival collection (arc. desc) or in any of the two fields (bio+arc. desc.), the relation between the two historians can be identified with maximum precision (p(1)=1). However, the recall is low (0.17 <= r(1) <= 0.25), and the accuracy too (0.67 <= a <= 0.69), due to many false negative results. Likewise, relations between historians' collections based on the mention in the biography can be predicted with high precision (p(1)=0.93) and average recall (r(1)=0.41).
- Relations between historians based on shared topics (topics) have lower precision (p(1)=0.65) and low recall (r(1)=0.18). Together with references in the archival description and/or biography, the precision increases (0.8 <= p(1) <= 0.9), while the recall is still low (0.17 <= r(1) <= 0.25). Surprisingly, relations between collections based on similar research topics cannot be detected by any of the proposed models, which are not able to predict any true positive result.

- When historians are connected via the same institutions (inst.), relations between them are predicted with low precision (p(1)=0.67) and high recall (r(1)=0.91). Again, relations between their collections are difficult to predict when based solely on shared institutions.
- Relations between historians sharing both topics and institutions (inst. + topics) are predicted with low precision (p(1)=0.68), and lower recall (r(1)=0.38). Relations between collections instead are predicted with high precision (p(1)=1) and rather low recall (r(1)=0.19).

For each combination of features (i.e. columns of prior tables), we selected the model that performed better in terms of precision, privileging the most recurring model in case of even results (i.e. Bayes). We analyse predicted results to understand whether these are known or unknown results. Specifically, in the EDA we demonstrated that relations between historians that are recorded in at least one biography correspond to valid relations in 100% of cases. Known relations are predicted relations for which a reference in the biography of one of the two historians is available. Unknown relations are those that do not appear in any biography. In Table 7 we show the percentage of known and unknown relations that each model returns for each combination of features.

Predicted relations between historians are mostly known when a reference in the biography or the archival description is used as a feature to predict relations. Predictions based on topics (77% unknown), institutions (55% unknown) or both (60%) provide more insights. Less insights are provided by relations between collections, where the highest number of unknown relations is provided by the number of institutions shared between historians (33%).

### 5. Discussion

Findings allow us to characterise art historical communities and to draw some conclusions on the broader applicability of our results.

The preliminary analysis confirmed the assumptions of experts in terms of relevant features. The EDA showed that relations between historians that are recorded in at least one biography, always refer to actual interactions between people. However, between 68% and 80% valid relations not recorded in any biography are still undisclosed.

**Relations between historians.** When historians are mentioned in another cataloguing record (regardless if in the biography or the archival description) the mention is likely to address actual interactions between people at hand. This claim can be generalised as a general attitude of cataloguers in recording mainly information relevant to the historian's working relations, rather than personal, unrelated information. In this case, *a priori* rules are sufficient to recommend relations to cataloguers.

Investigating shared topics and/or institutions revealed being the right direction to look into. The recall increases significantly (Table 5, r(bio)=0.19; r(topics)=0.18; r(inst)=0.91; r(topic+inst)=0.38) and the number of unknown relations that cannot be detected via *a priori* rules is significant (Table 7, topics=77%; institutions=55%; both=60%). To this extent, we believe probabilistic models can contribute with new insights when recommending relations.

However, the precision of predictions based on topics and institutions only, is lower than expected (0.65 <= p(1) <= 0.67), which requires further investigation. Institutional relations seem to offer promising insights compared to topic-based relations alone. According to annotators' comments, sharing broad topics does not imply a collaboration took place, since historians may focus on different aspects (e.g.

studying architecture or figurative arts of Renaissance). Instead, working in the same institutions seems to be a greater gluing factor.

**Relations between collections.** The EDA revealed that while historians mentioned in others' biographies are likely to have interacted, materials produced by historians that interacted with each other (e.g. correspondence, exchange of expertise on artworks) are not always likely to be preserved in other historians' collections. In fact, relations between collections are only valid 19% of times when historians interacted. To this extent, both probabilistic and *a priori* rules can be applied to recommend relations based on mentions in biographies with satisfying results ($p(1) = 0.93$, $r(1)=0.41$).

Although we expected collections catalogued according to similar research topics could be the result of historians' collaborations, such relations cannot be predicted, neither via deterministic rules nor via probabilistic models. Again, the combination of institutions and research topics seems to offer more insights (Table 7, 33% and 17%). Nonetheless, further investigation is required to improve results.

**Limitations.** Although ARTchives is a homogeneous (e.g. similar historians' life spans, research topics), highly curated, and representative source of the dynamics of a community (historians of Italian Modern Art), the size of the corpus of relations is still small to generalise results, which may not reflect general trends. In our dataset, significant support is provided for the following features: shared research topics, shared institutions, mention in the biography, mention in the archival collection. Instead, relations between historians based on the topic at hand, and relations between collections and other historians, are not sufficiently supported in the dataset, therefore they cannot be adequately analysed.

Using cross-validation to evaluate models was useful to cope with the small size of the dataset. However, despite we discarded less represented features, there is still an imbalance in the dataset used to predict relations between collections. We excluded from the analysis the direction of links between collections and historians, and we focused on bidirectional relations between collections, hence we cannot track all documents of a historian. We expect the insights we gathered in this preliminary study may vary when data will reach a critical mass. This aspect will deserve attention in future works since it is compelling for historiographic enquiries.

**Implications of the work.** Linked Open Data is massively used by archives in collaborative cataloguing projects, which suffer from data quality issues similar to those described here (Daquino, 2021b). Such systems can leverage results of our work to plan interventions to ensure data completeness, by either enforcing good practices in data collection - therefore ensuring a priori rules are sufficient to recommend new relations - or by leveraging our approach to discover, validate, and use features characterising their field for recommendation purposes.

To the best of our knowledge, this work is a pioneering attempt to characterise relations between historians in the art historiographic community with quantitative methods. While computational methods are differently applied to enquiries relevant to art historiography (e.g. image content classification, style detection, record linking), social aspects of the discipline are often overlooked, and computational methods have not been integrated across the discipline (Jaskot, 2020). This work is an initial attempt to combine symbolic and sub symbolic AI approaches to characterise and trace the history of art history.

## 6. Conclusion

In this article we have explored methods to characterise and predict relations extracted from archival data that are relevant to historiography of art. Results of our preliminary work show limited but promising insights. We were able to characterize relations between historians and relations between their collections as driven by a priori rules when partial information is provided in cataloguing records. To this extent, we can confidently claim that archival recommender systems do not need to rely on machine learning techniques when suggesting data to cataloguers if their priority is the precision of recommendations. Nonetheless, if the cataloguing process suffers from data completeness, most potential recommendations will remain unveiled. We demonstrated that discovering relations between historians is a suitable task for classification models when institutional relations and topic-based relations can be extracted. While topic-based relations are hard to detect, institutional relations and the combination of institutions and topics can be helpful. Our results show a lower precision than desired. Nonetheless, predictions include most of the unknown relations which would make a significant impact in the recommendation system together with *a priori* rules.

Future work will focus on better framing relations based on topics, institutions, and topics/institutions. Other data integration strategies will be used to fill the gaps in the dataset and to refine the classifier. Historians' network analysis, including historians mentioned in biographies but not recorded in ARTchives, will be addressed so as to consider different features and improve results.

Acknowledgments
We would like to thank the ARTchives consortium that is actively promoting the data crowdsourcing campaign to build the knowledge graph of art historians' archival collections.

# 7. Notes

[1] Authors' responsibility: Lucia Giagnolini is responsible for writing sections 3.3, 3.4, 4.1; Marilena Daquino is responsible for writing section 1, 2, 3.1, 3.2, 4.2; all authors are responsible for writing section 5,6.

[2] http://artchives.fondazionezeri.unibo.it

[3] The catalogue of the Federico Zeri's photo archive is available at: http://catalogo.fondazionezeri.unibo.it/search/work

[4] ARTchives SPARQL endpoint, https://artchives.fondazionezeri.unibo.it/sparql

[5] DBpedia spotlight service, https://www.dbpedia-spotlight.org/

[6] SpaCy python library, https://spacy.io/

[7] http://artchives.fondazionezeri.unibo.it/collection-1598630286-3009102

[8] The jupyter notebook including the preliminary EDA is available at https://github.com/marilenadaquino/ARTchives/blob/master/data_analysis/EDA_ARTchives.ipynb

[9] The jupyter notebook including the EDA is available at: https://github.com/marilenadaquino/ARTchives/blob/master/data_analysis/EDA_ARTchives.ipynb

[10] The jupyter notebook is available at: https://github.com/marilenadaquino/ARTchives/blob/master/data_analysis/ARTchives_classifier.ipynb

# Tables

**Table 1** Overview of fields available in the dataset and extraction/annotation methods

| Table | # | Column | Method | Source |
|---|---|---|---|---|
| artists_periods | A1 | art_hist_1 | Graph extraction | |
| | A2 | art_hist_2 | Graph extraction, NER | biography |
| | A3 | artist_or_period | Graph extraction, NER | biography |
| | A4 | Does any relation exist between the two historians? | Manual annotation | - |
| | A5 | Is the relation recorded in at least one historian biography available on ARTchives? | Graph extraction, NER | biography |
| | A6 | Is the relation recorded in both the historians' biographies available on ARTchives? | Graph extraction, NER | biography |
| | A7 | Did they collaborate / interact on the period/artist at hand? | Manual annotation | - |
| | A8 | Is historian 2 relevant to historian 1's archive? | Manual annotation | - |
| | A9 | Is historian 2 mentioned in historian 1's archive on ARTchives? | Graph extraction, NER | Collection |
| | A10 | Is historian 1 relevant to historian 2's archive? | Manual annotation | - |
| | A11 | Is historian 1 mentioned in historian 2's archive on ARTchives? | Graph extraction, NER | Collection |
| institutions | I1 | art_hist_1 | Graph extraction | |
| | I2 | art_hist_2 | Graph extraction, NER | biography |
| | I3 | institution | Graph extraction, NER | biography |
| | I4 | Does any relation exist between the two historians? | Manual annotation | - |
| | I5 | Which one? If known | Manual annotation | - |

**Table 2 Overview of relations relevant to art historiography**

| #  | Subject    | Predicate                   | Object     | Status    | Validation          |
|----|------------|-----------------------------|------------|-----------|---------------------|
| 1  | Historian  | produced                    | Collection | available |                     |
| 2  | Historian  | subject                     | Topic      | available |                     |
| 3  | Historian  | subject                     | Institution| available |                     |
| 4  | Historian  | subject                     | Historian  | available |                     |
| 5  | Historian  | Interacted with             | Historian  | new       | #4, annot. #A4, #I4 |
| 6  | Historian  | Interacted on               | Topic      | new       | #5, annot., #A7     |
| 7  | Collection | Produced by                 | Historian  | available |                     |
| 8  | Collection | subject                     | Historian  | available |                     |
| 9  | Collection | Includes materials relevant to | Historian | new    | #8, annot. #A8, #A10|
| 10 | Collection | subject                     | Topic      | available |                     |
| 11 | Collection | held by                     | Institution| available |                     |
| 12 | Collection | subject                     | Institution| available |                     |
| 13 | Collection | is related to               | Collection | new       | Annot. #A8, #A10    |

**Table 3 EDA results regarding relations between historians**

|  | 1. Total relations between historians | 2. Total unique pairs of historians | 3. Valid relations between historians | 4. Valid unique pairs of historians | 5. Valid relations not recorded in ARTchives | 6. Valid relations based on the shared topic |
|---|---|---|---|---|---|---|
| 1. At least one topic in common | 332 | 173 | 162 | 71 | 124 | 52 |
| 2. At least one institution in common | 60 | 49 | 39 | 33 | 29 | n/a |
| 3. At least one institution and one topic in common | 173 | 32 | 71 | 22 | 57 | 5 |

**Table 4 EDA results regarding relations between historians and collections**

|  | 1. Historian-collection relations | 2. Historian-collection relations not recorded in ARTchives biographies |
|---|---|---|
| 1. At least one topic in common | 31 | 19 |
| 2. At least one institution and one topic in common | 13 | 4 |

**Table 5 Evaluation of classification models for relations between historians**

|    | bio | arch. desc | bio + arc.desc. | topics | topics + bio | topics + arc.desc. | topics + bio + arc. desc. | inst. | Inst. + topics |
|----|-----|-----------|-----------------|--------|--------------|--------------------|---------------------------|-------|----------------|
| lr | **p=0.8 p(1)=1 r(1)=0.17 a=0.67** | **p=0.8 p(1)=1 r(1)=0.17 a=0.66** | **p=0.9 p(1)=1 r(1)=0.25 a=0.69** | p=0.59 p(1)=0.59 r(1)=0.32 a=0.63 | p=0.83 p(1)=0.78 r(1)=0.35 a=0.69 | p=0.83 p(1)=0.77 r(1)=0.32 a=0.68 | p=0.85 p(1)=0.79 r(1)=0.38 a=0.70 | p=0.64 p(1)=0.66 r(1)=0.94 a=0.78 | p=0.7 p(1)=0.69 r(1)=0.38 a=0.67 |
| nb | **p=0.8 p(1)=1 r(1)=0.19 a=0.67** | **p=0.8 p(1)=1 r(1)=0.17 a=0.66** | **p=0.9 p(1)=1 r(1)=0.25 a=0.69** | p=0.57 p(1)=0.61 r(1)=0.26 a=0.63 | **p=0.8 p(1)=1 r(1)=0.19 a=0.67** | **p=0.8 p(1)=1 r(1)=0.17 a=0.66** | **p=0.9 p(1)=1 r(1)=0.25 a=0.69** | **p=0.65 p(1)=0.67 r(1)=0.91 a=0.77** | p=0.61 p(1)=0.68 r(1)=0.24 a=0.64 |
| dt | **p=0.8 p(1)=1 r(1)=0.19 a=0.67** | p=0.7 p(1)=1 r(1)=0.17 a=0.66 | **p=0.9 p(1)=1 r(1)=0.25 a=0.69** | **p=0.57 p(1)=0.65 r(1)=0.18 a=0.62** | p=0.73 p(1)=0.75 r(1)=0.29 a=0.67 | p=0.63 p(1)=0.74 r(1)=0.28 a=0.66 | p=0.78 p(1)=0.81 r(1)=0.29 a=0.68 | p=0.63 p(1)=0.65 r(1)=0.91 a=0.62 | **p=0.72 p(1)=0.68 r(1)=0.38 a=0.67** |

**Table 6** Evaluation of classification models for relations between historians' collections

|    | bio | topics | bio + topics | inst. | Inst. + topics |
|----|-----|--------|--------------|-------|----------------|
| lr | **p=0.76**<br>**p(1)=0.93**<br>**r(1)=0.41**<br>**a=0.89** | p=0<br>p(1)=0<br>r(1)=0<br>a=0.82 | **p=0.76**<br>**p(1)=0.93**<br>**r(1)=0.41**<br>**a=0.89** | **p=0.2**<br>**p(1)=0.6**<br>**r(1)=0.09**<br>**a=0.82** | p=0.2<br>p(1)=0.6<br>r(1)=0.09<br>a=0.82 |
| nb | **p=0.76**<br>**p(1)=0.93**<br>**r(1)=0.41**<br>**a=0.89** | p=0<br>p(1)=0<br>r(1)=0<br>a=0.79 | **p=0.76**<br>**p(1)=0.93**<br>**r(1)=0.41**<br>**a=0.89** | p=0.16<br>p(1)=0.33<br>r(1)=0.12<br>a=0.8 | p=0.32<br>p(1)=0.42<br>r(1)=0.16<br>a=0.81 |
| dt | **p=0.76**<br>**p(1)=0.93**<br>**r(1)=0.41**<br>**a=0.89** | p=0<br>p(1)=0<br>r(1)=0<br>a=0.82 | p=0.76<br>p(1)=0.92<br>r(1)=0.38<br>a=0.88 | p=0<br>p(1)=0<br>r(1)=0<br>a=0.81 | **p=0.5**<br>**p(1)=1**<br>**r(1)=0.19**<br>**a=0.85** |

**Table 7 Percentage of known/unknown predicted relations**

| Relations between historians | | | |
|---|---|---|---|
| Feature | Model | Known (%) | Unknown (%) |
| Bio | nb | 100 | 0 |
| Arch. desc | nb | 67 | 33 |
| Bio + arch. desc. | nb | 77 | 23 |
| Topics | dt | 23 | 77 |
| Topics + bio | nb | 77 | 23 |
| Topics + arch. desc. | nb | 67 | 33 |
| Topics + arch. Desc. + bio | nb | 76 | 23 |
| Inst. | nb | 45 | 55 |
| Inst. + topics | dt | 40 | 60 |
| Relations between historians' collections | | | |
| Bio | dt | 100 | 0 |
| Topics | n/a | n/a | n/a |
| Topics + bio | nb | 100 | 0 |
| Inst. | lr | 67 | 33 |
| Topics + inst. | dt | 83 | 17 |

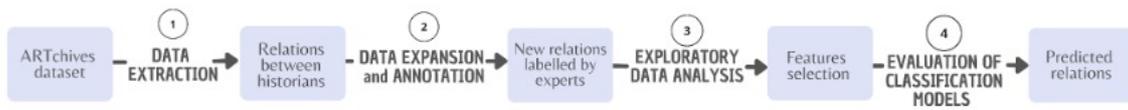

Fig. 1 Overview of our approach

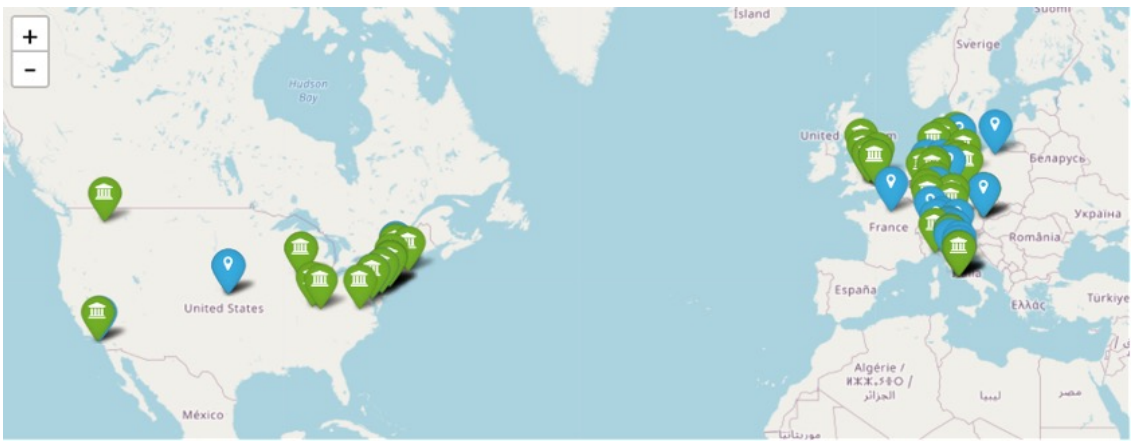

Fig. 2 The network of art historians based on their research topics

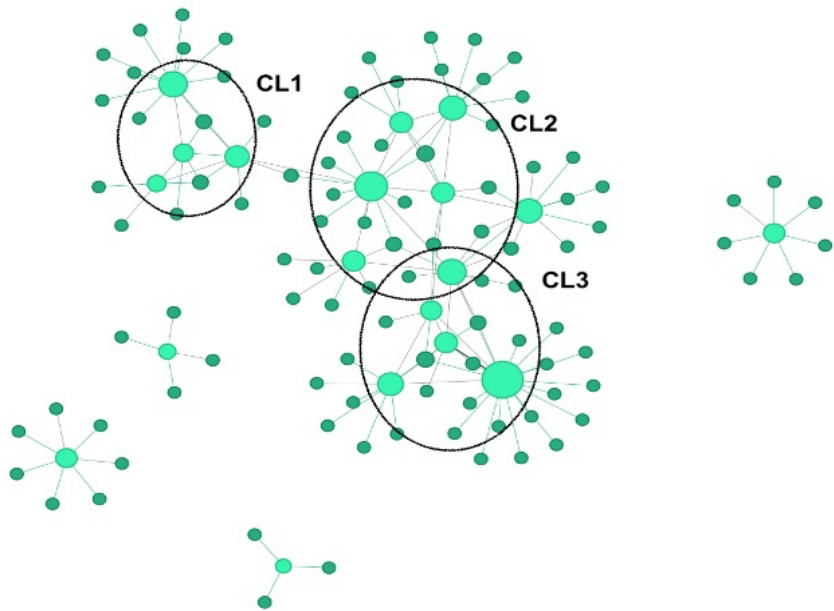

Fig. 3 Distribution of art historians' places of education and work

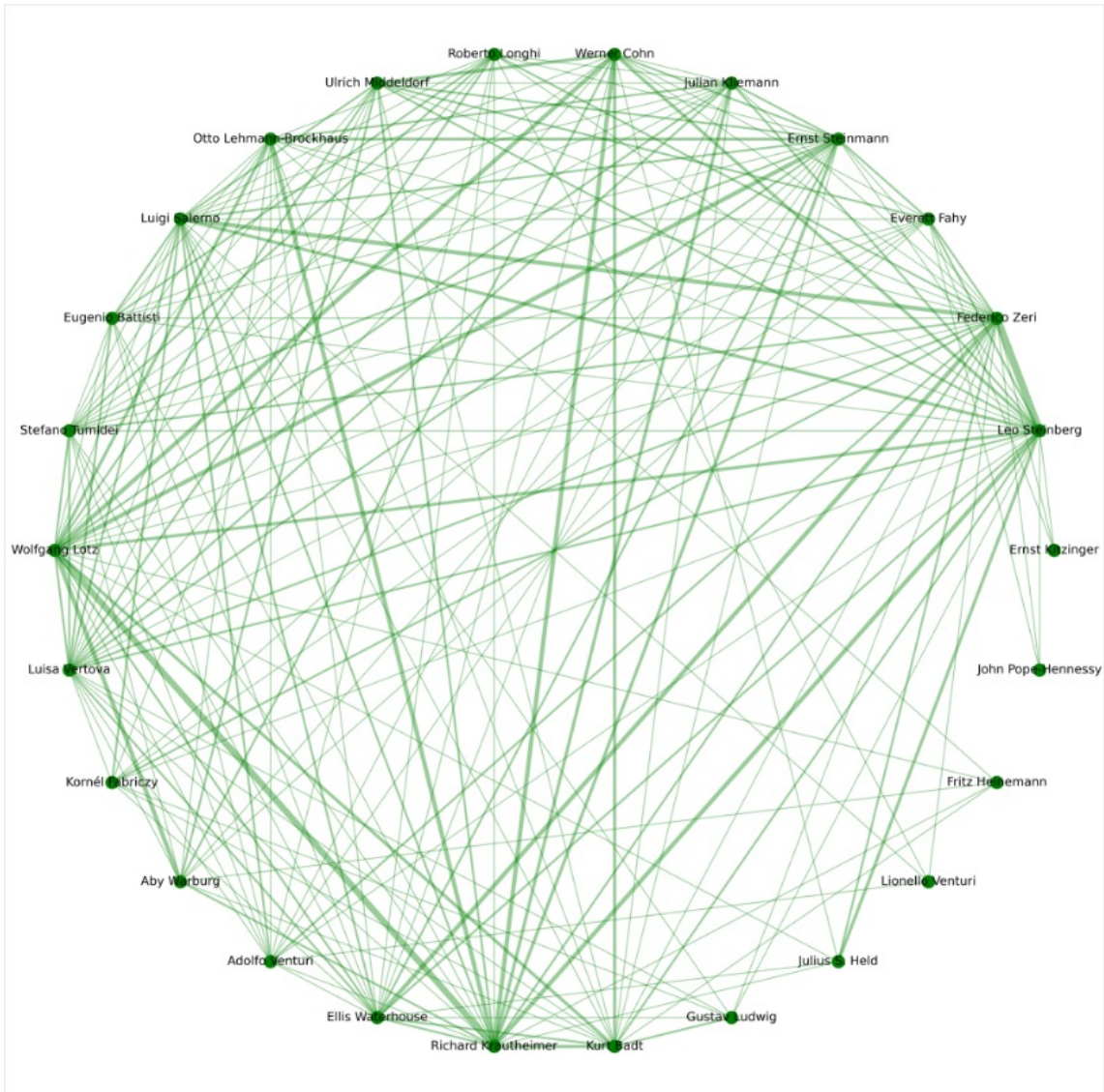

Fig. 4 Graph network of art historians based on the common education or workplaces